# OMG-Net: A Deep Learning Framework Deploying Segment Anything to Detect Pan-Cancer Mitotic Figures from Haematoxylin and Eosin-Stained Slides

Zhuoyan Shen[1*], Mikael Simard[1], Douglas Brand[1, 2], Vanghelita Andrei[3, 4], Ali Al-Khader[3, 4], Fatine Oumlil[4], Katherine Trevers[3, 4], Thomas Butters[3], Simon Haefliger[3,5], Eleanna Kara[6], Fernanda Amary[3, 4], Roberto Tirabosco[3, 4], Paul Cool[7,8], Gary Royle[1], Maria A. Hawkins[1,2], Adrienne M. Flanagan[†3,4], Charles-Antoine Collins Fekete[1†*]

1. Department of Medical Physics and Biomedical Engineering, University College London, United Kingdom.
2. Department of Radiotherapy, University College London Hospitals NHS Foundation Trust, United Kingdom.
3. Research Department of Pathology, University College London Cancer Institute, United Kingdom.
4. Cellular and Molecular Pathology, Royal National Orthopaedic Hospital NHS Foundation Trust, United Kingdom.
5. Institute of Medical Genetics and Pathology, University Hospital Basel, University of Basel, Basel, CH, Switzerland.
6. Department of Neurology, Rutgers Biomedical and Health Sciences, Rutgers University, New Jersey, United States.
7. Department of Orthopaedics, The Robert Jones and Agnes Hunt Orthopaedic Hospital, United Kingdom.
8. School of Medicine, Keele University, United Kingdom.

†: Both authors contributed equally to this work.

*: Corresponding authors

## Abstract

Mitotic activity is an important feature for grading several cancer types. Counting mitotic figures (MFs) is a time-consuming, laborious task prone to inter-observer variation. Inaccurate recognition of MFs can lead to incorrect grading and hence potential suboptimal treatment. In this study, we propose an artificial intelligence (AI)-aided approach to detect MFs in digitised haematoxylin and eosin-stained whole slide images (WSIs). Advances in this area are hampered by the limited number and types of cancer datasets of MFs. Here we establish the largest pan-cancer dataset of mitotic figures by combining an in-house dataset of soft tissue tumours (STMF) with five open-source mitotic datasets comprising multiple human cancers and canine specimens (ICPR, TUPAC, CCMCT, CMC and MIDOG++). This new dataset identifies 74,620 MFs and 105,538 mitotic-like figures. We then employed a two-stage framework (the Optimised Mitoses Generator Network (OMG-Net) to classify MFs. The framework first deploys the Segment Anything Model (SAM) to automate the contouring of MFs and surrounding objects. An adapted ResNet18 is subsequently trained to classify MFs. OMG-Net reaches an F1-score of 0.84 on pan-cancer MF detection (breast carcinoma, neuroendocrine tumour and melanoma), largely outperforming the previous state-of-the-art MIDOG++ benchmark model on its hold-out testing set (*e.g.* +16% F1-score on breast cancer detection, $p<0.001$) thereby providing superior accuracy in detecting MFs on various types of tumours obtained with different scanners.

# Highlights

- A large pan-cancer mitotic figure dataset has been created by enhancing open-source datasets and integrating an in-house dataset for soft tissue tumours across 20 subtypes.
- A novel nuclei detection framework, based on Segment Anything, has been developed, demonstrating state-of-the-art performance in detecting pan-cancer mitotic figures.
- The results demonstrate that incorporating the contours of nuclei significantly enhances the accuracy and robustness of mitotic figure detection.
- The feasibility of zero-shot deployment of foundation models for data generation, standardization, and nuclei detection model development is demonstrated and discussed.

# Introduction

Mitotic activity is a crucial indicator of cellular proliferation and plays a pivotal role in cancer diagnosis and guiding clinical management (Williams & Stoeber, 2012). Counting mitotic figures (MFs) from haematoxylin and eosin (H&E)-stained whole slide images (WSIs) is a fundamental task in pathology, required for the grading of some tumours. By convention, in clinical practice, mitotic counts are performed in the 10 most mitotically active high-power microscopic fields (HPFs) within a tumour (Cree, et al., 2021). As this is a time-consuming task, and subject to significant inter-observer variability (Malon, et al., 2012; Veta, et al., 2016; Robbins, et al., 1995), there has been considerable interest and effort in the development of automated MF detection models, *e.g.* ICPR (Capron & Genestie, 2011; Roux, 2014) and TUPAC (Veta, et al., 2019) initiated the development of breast cancer MF datasets. Initially, mitotic detection models focused on learning handcrafted features (Irshad, 2013; Tashk, et al., 2013; Paul, et al., 2015), but recently transitioned to deep-learning-based methods that show promise (Mahmood, et al., 2020; Sebai, et al., 2020; Li, et al., 2018; Cai, et al., 2019). However, MF detection remains a challenging task (Aubreville, et al., 2023), due to the different appearance of MF in the four phases of mitosis, the range of features exhibited by abnormal MFs, as well as structures that mimic MFs (mitotic-like figures, MLFs). The above challenges are compounded by the histological heterogeneity in normal tissues and tumour types, staining variation between labs and differences in digital scanners used to generate WSIs.

To improve the detection of MF, the MItosis DOmain Generalization (MIDOG) (Aubreville, et al., 2021; Aubreville, et al., 2020) published an updated version of their multi-domain dataset, MIDOG++ (Aubreville, et al., 2023). This contains 503 annotated images across seven different cancer types, representing the largest currently available published dataset of MFs. The data utilised in the MIDOG studies contains the HPFs manually selected by pathologists to mimic clinical practice. However, the pathologist-led decisions may not be reproducible because of the recognised inter-observer variation (Meyer, et al., 2005; Bertram, et al., 2019), and discrepancies can be caused by the selection of areas with the densest mitotic activity (Diest, et al., 1992). In contrast, the CMC (Aubreville, et al., 2020) and CCMCT (Bertram, et al., 2021) datasets used AI-assisted annotations to generate large-scale WSI datasets for MFs using canine cancers. The former used 21 WSIs of canine mammary carcinomas whereas the CCMCT dataset included 32 WSIs of canine mast cell tumours. These studies demonstrated that annotating MFs on a WSI improves the robustness of classifiers by removing the HPF selection bias and leads to a significantly higher number of detected mitoses, helping to refine further training (Bertram, et al., 2021).

One preferred approach to take forward this field of MF detection would have been to increase the size of the existing datasets incorporating multiple scanner types, staining differences across multiple sites, and tumour types. However, the lack of standardisation in the annotation protocol across existing various datasets limit their integration. For example, in the ICPR, each pixel within MFs was labelled, whereas the TUPAC only encircled MFs. MIDOG++, CCMCT, and CMC utilised bounding boxes to denote the targets. We therefore took the approach to standardise the annotations by contouring nuclei of MFs.

Historically, targets in cellular object detection tasks are denoted using bounding boxes. However, several studies have reported that incorporating a target's mask facilitates model training and improves the overall classification performance. For instance, the Mask-RCNN outperformed the Faster-RCNN in a variety of object detection tasks (He, et al., 2018), including MF detection (Sebai, et al., 2020). The advantages of integrating nuclei contours for detection include enhancing the definition of nuclei boundaries, mitigating the morphological variability of the MFs (Li, 2023) and reducing the impact of tumour histological heterogeneity. Given the constraints of a small dataset and the significant variability between mitotic cells, introducing a recognisable mitotic feature into the model aids in stabilising the training process and leads to a faster convergence.

The aim of this study was to improve the detection of MF across multiple tumour types. First, we established a large uniform database of pan-cancer MFs by deploying the Segment Anything Model (SAM) (Kirillov, et al., 2023), a foundation object detection model, in five open-source datasets (ICPR, TUPAC, CCMCT, CMC, MIDOG++) using a single nuclei mask format. Manual revision of the masks was performed to maximise database quality. Then, we contributed an in-house dataset of human soft tissue tumours (STT) MFs (N=8,400) (Soft-Tissue Mitotic Figures, STMF). Although STT represents a rare tumour group, they comprise over 100 subtypes exhibiting a wide variety of histological appearances and mimic other tumours including common cancers such as melanoma, carcinoma and lymphoma. STT harbours a variable number of MFs and aids in reaching a diagnosis and predicting disease behaviour (Coindre, 2006). As of now, no publicly accessible data have been published for MFs in STT. The STMF was initiated by staining WSIs with an anti-phosphorylated histone H3 (pHH3) antibody to target MFs which was expanded and improved by AI-assisted annotations made by pathologists.

The second objective was to develop an improved MF detection framework, which we named Optimised Mitoses Generator Network (OMG-Net). By integrating nuclei masks into the pre-trained classifier via a first-layer addition, we allow the model to focus on the morphological features of MFs. We demonstrate that OMG-Net is both more sensitive and specific at detecting objects, including MFs, throughout the input WSI, compared to previous models.

# Material and methods

## *Dataset*

**Figure 1** illustrates the data generation pipeline for the in-house dataset, STMF, and the curation process for the multi-source datasets including the STMF and the open-source datasets.

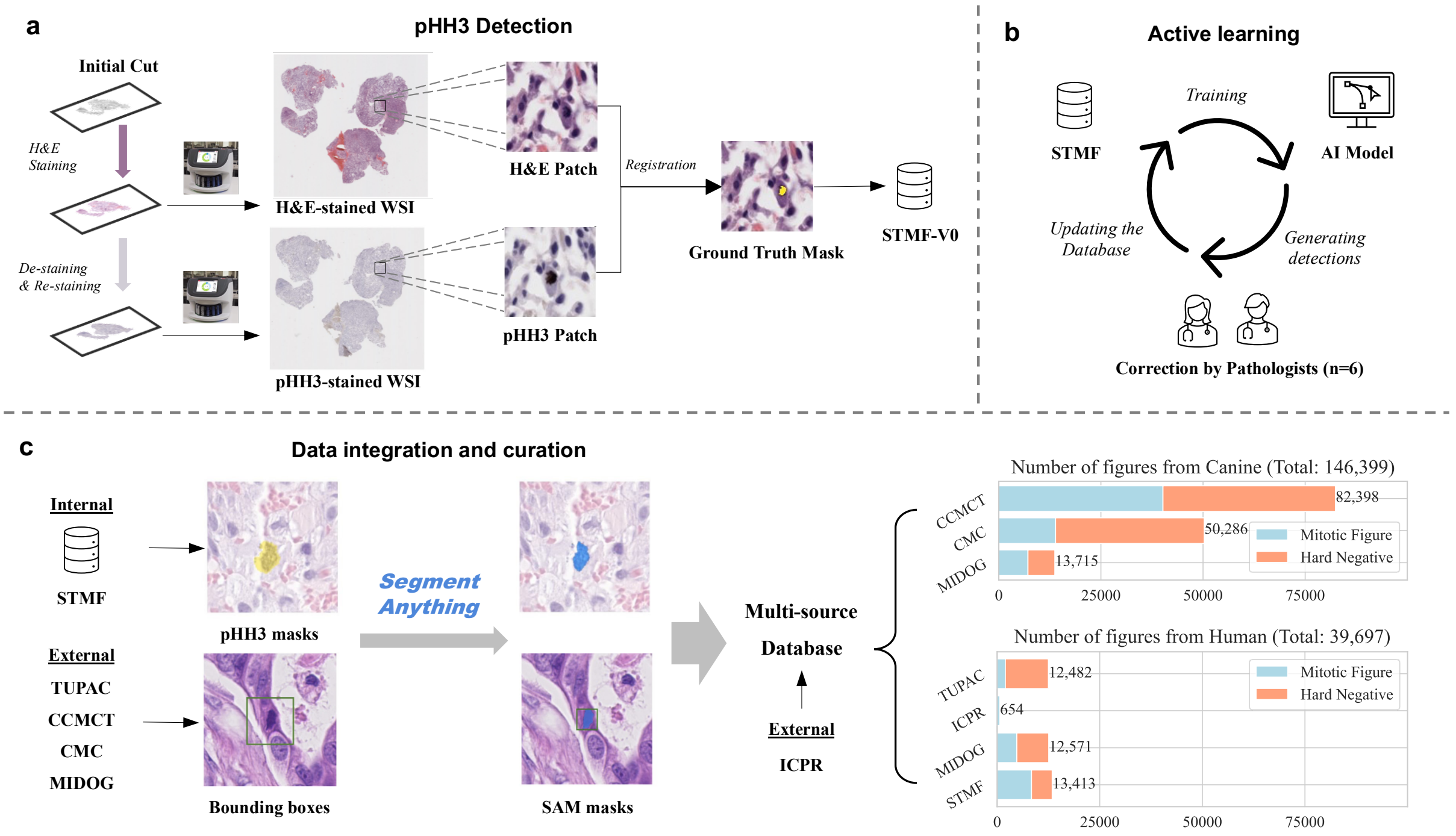

**Figure 1: Data preparation workflow. a** Haematoxylin and eosin (H&E)-stained whole slide images (WSIs) were de-stained after which immunohistochemistry was performed using an anti-phosphorylated histone H3 (pHH3) antibody which labels mitotic figures(MFs) (STMF-V0). **b** An initial Mask-RCNN model trained on STMF-V0 was applied to new WSIs for detecting MFs, which were then labelled by six pathologists as MF or false positives. This process facilitated the iterative refinement and expansion of the dataset to produce STMF. **c** The masks of the MFs from STMF and the bounding boxes from four external datasets were refined by Segment Anything (SAM) and integrated with ICPR to create the final dataset.

### *Open-source datasets*

We integrated five open-source datasets (ICPR, TUPAC, CCMCT, CMC, MIDOG++), comprising 68,687 MFs from eight different scanners and eight types of human and canine tumours. The types of tumours studied and scanners are listed in **Supplementary Table 1**. All the images were scanned in 40× magnification with a pixel size of approximately 0.25 µm.

### *In-house dataset*

We describe a workflow for utilising an anti-phosphorylated histone H3 (pHH3) antibody to specifically detect MFs and expand the dataset by active learning (**Figure 1**). The number of MFs in each diagnosis of soft tissue tumours is listed in **Supplementary Table 2.**

- pHH3-assisted MF detection: the pHH3 antibody employed specifically detects the core protein histone H3 only when phosphorylated at serine 10 (Ser10) or serine 28 (Ser28), thereby identifying mitotic cells within a tissue sample (Elmaci, et al., 2018). We selected 94 archived slides and tissue blocks from soft tissue tumours and prepared fresh H&E tissue sections which were then scanned for generating our dataset. These H&E-stained tissue sections were then de-stained after which immunohistochemistry was performed using a rabbit monoclonal (RM) hybridoma Ser10 pHH3 [BC37] (Tacha, 2015) and then counterstained with eosin. The masks of the MFs were extracted from pHH3-immunolabelled WSIs by setting thresholds for the RGB values and transferred to the same location on the matching H&E-stained WSIs. Registration between the pHH3-immunolabelled and H&E-stained WSIs was achieved by random sample consensus (RANSAC) (Fischler & Bolles, 1981) on both a WSI-level and patch-level. The contours and positions of 7,952 MFs (STMF-V0), were identified and validated by pathologists reviewing the H&E and immunolabelled sections. However, not all mitoses were identified by pHH3-labelling indicating that the antibody was not entirely sensitive (Ribalta, et al., 2004).
- Active Learning: Although the identification of cells in mitosis by pHH3 can establish a dataset with a large number of MFs, it cannot identify MLFs, and models trained only with IHC suffer from limited precision. Active learning is required to augment the dataset with MLFs.

  During the active learning process, pathologists corrected the image labels given by a machine learning model and fed them back to re-train the initial model, so that the model performance for the target task can be continuously improved during the iteration of machine-generating and human-labelling.

  To expand the STMF-V0 dataset, we trained an initial Mask-RCNN model on it and applied the model to new WSIs for detecting MFs. The AI-detected MFs were randomly assigned to six pathologists to be independently labelled as 'MF', 'not MF' if the pathologist could confidently make a decision, or 'uncertain' when the morphological features were equivocal. These equivocal MFs were reviewed by two senior pathologists. Other structures such as apoptotic bodies were also labelled to create the final dataset, STMF, with 8400 MFs and 5035 MLFs.

*Ethical Approvals*

The data involved in the STMF dataset are collected in the Royal National Orthopaedic Hospital (RNOH) NHS Trust under the Health Research Authority (HRA) and Health and Care Research Wales (HCRW) Approval. Integrated Research Application System (IRAS) project ID: 328987. Protocol number: EDGE 161548. Research Ethics Committee (REC) reference: 23/NI/0166. Informed consent was obtained from all human participants.

***Data Curation***

The MFs were annotated using bounding boxes in the CCMCT, CMC and MIDOG++ datasets. However, the size of the boxes varies due to the lack of standard annotation criteria. We hypothesised that the contours of nuclei could provide extra information for classifying MFs, as the model would be guided to focus on the most representative pixels of the nuclei rather than the surrounding environment.

We use the bounding boxes provided in the CCMCT, CMC and MIDOG++ datasets as prompts to generate the masks using SAM. To ensure the quality of the automatically generated masks, we inspected individual masks of the MFs from three types of human tumours and canine soft tissue sarcoma in MIDOG++. The percentage of masks amended following review is 8%, 5% and 16% out of 4435 masks in breast carcinoma, 2075 in melanoma and 2400 in neuroendocrine tumour, respectively. In total, only 8% of the masks required a second inference of SAM using adjusted bounding boxes. Since the cells can be distorted during the de-staining and pHH3 labelling process, we also applied the SAM to the STMF using the outside boxes of the pHH3-immunolabelled masks as prompts. The numbers of MFs and MLFs from human and canine samples are shown in **Figure 1**. Quality assurance was done for masks of all the human samples, whereas the generation of masks in canine sections was fully automated.

### ***OMG-Net: A Two-Stage Detection Framework***

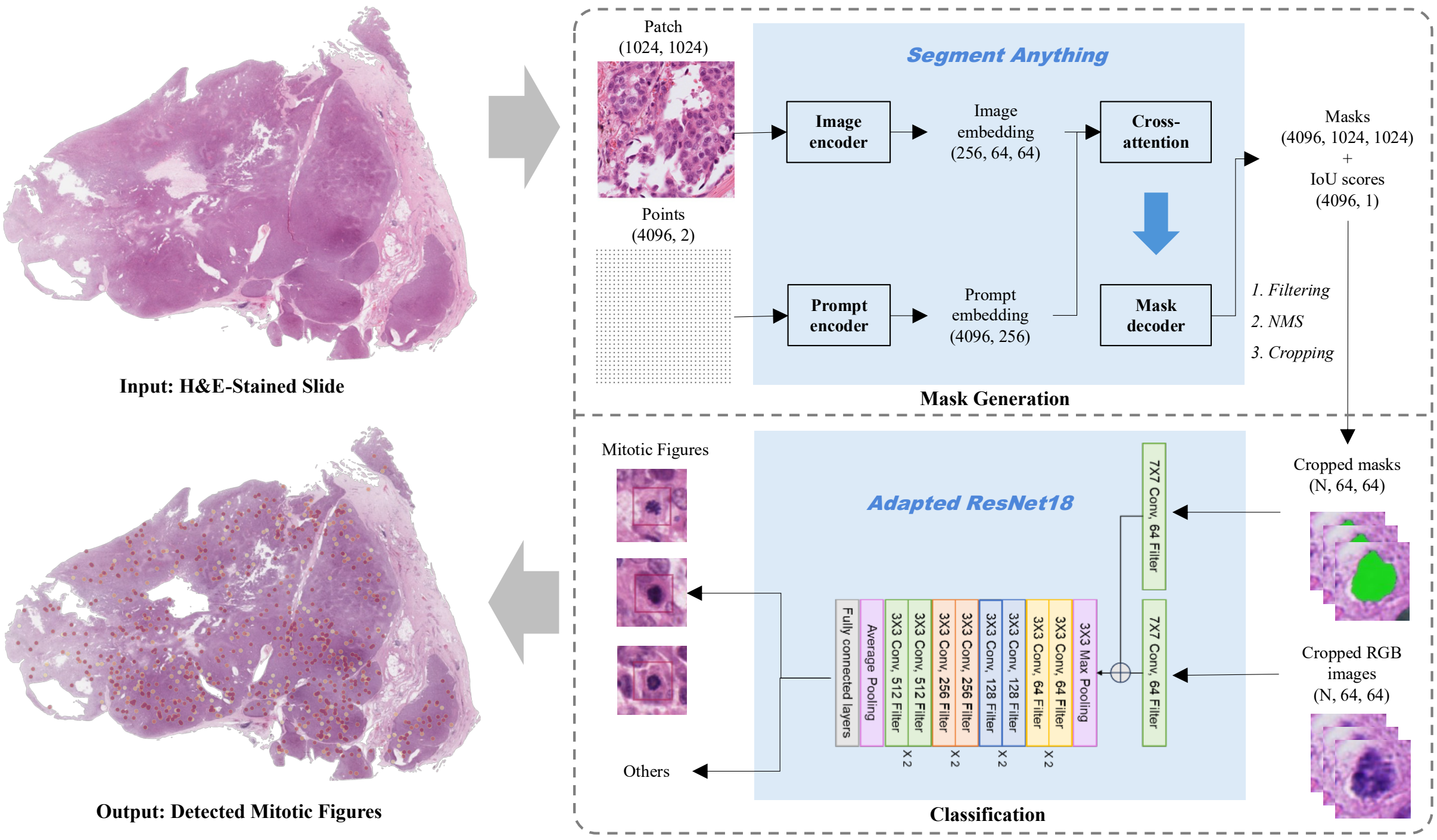

**Figure 2: The architecture of the OMG-Net.** The two-step architecture includes mask generation and mitotic figures (MF) classification. First, the post-process cell masks from patched WSIs are generated by Segment Anything (SAM) using an evenly sampled point grid as a prompt. Second, the RGB image of the segmented cell and the binary mask are used to classify MFs by employing an adapted ResNet18.

The structure of OMG-Net is outlined in **Figure 2**. The proposed framework consists of two steps:

- The SAM was applied to patches of 1024 × 1024 pixels from the tumour regions. 64 points were evenly sampled along each dimension, totalling 4,096 points used as prompts per patch. The quality of the masks was predicted by two factors, an AI-predicted Intersection over Union (AI-IoU) and a stability score. The AI-predicted IoU comes from an adjacent multi-layer perceptron in the mask decoder section of SAM. The stability score is the IoU between the binary masks obtained by thresholding the predicted mask logits at high and low values. Only the objects with AI-IoU scores and stability scores higher than 0.8 and areas between 2.25 μm$^2$ and 225 μm$^2$ were kept after filtering. The filtered masks were then ranked by their AI-IoU scores. Non-maximum suppression (NMS) (Hosang, et al., 2017)was used to remove duplicated masks.

- The objects generated were then classified by the second model, a ResNet18 pre-trained on ImageNet, as MFs or other objects. In addition to taking a 3-channel RGB image, the mask of the object was encoded by a convolutional layer and summed to the first convolutional layer of the ResNet18. Via this process, we retained the ability to use pre-trained models while providing extra mask information to the model.

***Model Development and Testing***

The framework was implemented using Pytorch and Pytorch Lightning and was trained using a single NVIDIA GeForce RTX 3090 for 30 epochs with a batch size of 8,000. The learning rate was set up at 0.001, optimised by the AdamW algorithm (Loshchilov & Hutter, 2019) and cosine annealing scheduler (Loshchilov & Hutter, 2017).

*Training and Validation*

We trained the ResNet18 to classify MFs while the SAM mask generator was not retrained. The SAM was applied to all patches in the dataset after data curation, and the other objects surrounding the targets were also segmented and included in the training and validation data. The binary classifier is trained on two classes: (1) MFs and (2) labelled MLFs and other cells or objects segmented by SAM. In each training process, 90% of the data was used for training the model, while the remaining 10% was used for validation. The training was repeated five times using different random seeds to get five models with different data splits.

*Data Augmentation*

Colour and spatial augmentation were applied to the training data to reduce the impact of the staining variation and increase the robustness of the model. To achieve colour augmentation, RGB images are deconvolved into H&E stains using the stain vectors proposed by Ruifrok and Johnston (Ruifrok & Johnston, 2001). The stain concentration perturbation scheme introduced by Tellez et al. (Tellez, et al., 2018) was used with a uniform sampling and $\sigma = 0.14$ on the deconvolved H&E channels prior to reconstructing RGB images. Random horizontal flips ($p = 0.4$) was also used.

*Test set and performance metrics*

We used the same testing set provided by MIDOG++, which contains 674 MFs from 56 sections of three types of human tumours. Precision, recall and F1 score were used to evaluate the performance of our mitotic detection framework. They were calculated by

$$Precision = \frac{N_{TP}}{N_{TP} + N_{FP}}$$

$$Recall = \frac{N_{TP}}{N_{TP} + N_{FN}}$$

$$F1 = 2 \cdot \frac{Precision \cdot Recall}{Precision + Recall}$$

where $N_{TP}$, $N_{FP}$ and $N_{FN}$ represent the number of true positives, false positives and false negatives, respectively. Mann–Whitney U test (Nachar, 2008) was used for comparing the unpaired scores of different models.

# Results

### *Developing a large-scale MF dataset*

We established a large in-house dataset for MFs in STT and merged it with five open-source datasets for MFs from human and canine specimens (**Table 1**). The final dataset contains 74,620 MFs and 105,538 MLFs from 712 different images or WSIs with the SAM-delineated masks for nuclei. Masks of human MFs were reviewed and modified to ensure the quality of nuclei contours. Additionally, the dataset included a large number of SAM-segmented objects, comprising tumour cells, immune cells, red blood cells, artefacts and any objects at the cell scale, collected during the data curation.

**Table 1. Number of different types of objects in the integrated dataset.**

| Dataset | Tumour Types | Number of Images | Number of MFs | Number of MLFs | Non-MF objects |
|---|---|---|---|---|---|
| ICPR | Breast carcinoma | 100 | 654 | 0 | 10,696 |
| TUPAC | Breast carcinoma | 73 | 1,999 | 10,483 | 233,992 |
| MIDOG++ | Breast carcinoma<br>Lung carcinoma*<br>Lymphosarcoma*<br>Neuroendocrine tumour*<br>Mast cell tumour*<br>Melanoma<br>Soft tissue sarcoma* | 392 | 9,470 | 11,433 | 559,827 |
| CMC | Breast carcinoma* | 21† | 13,907 | 36,379 | 2,428,456 |
| CCMCT | Mast cell tumour * | 32† | 40,190 | 42,208 | 1,082,776 |
| STMF | Soft tissue tumour | 103[1]† + 226[2] | 8,400 | 5,035 | 395,670 |
| Total | | 938 | 74,620 | 105,538 | 4,701,417 |

*Canine Specimens. †WSIs rather than selected regions.

[1] pHH3-immunohistochemistry was used for identifying MFs.

[2] Active learning was used for annotating MFs.

### *Performance of MF detection in various human tumours*

We benchmarked our OMG-Net against the current state-of-the-art MIDOG++ dataset/model containing three types of human tumours (breast carcinoma, neuroendocrine tumour, and melanoma). **Table 2** shows the mean precision, recall and F1 scores with the standard deviation of the proposed framework trained five times using different random seeds, along with the F1 score obtained by ensemble voting and the F1 scores quoted in the MIDOG++ paper.

**Table 2. Precision, recall and F1 scores in MIDOG++ testing set of OMG-Net against the model presented by MIDOG++**

| Tumour Types | Precision | Recall | F1 | Ensemble F1 | F1(MIDOG++) |
|---|---|---|---|---|---|
| Breast carcinoma | 0.82 ± 0.02 | 0.88 ± 0.02 | 0.85 ± 0.02 | **0.87** | 0.71 ± 0.02 |
| Neuroendocrine tumour | 0.64 ± 0.02 | 0.65 ± 0.03 | 0.64 ± 0.02 | **0.67** | 0.59 ± 0.02 |
| Melanoma | 0.83 ± 0.02 | 0.84 ± 0.03 | 0.83 ± 0.01 | **0.85** | 0.81 ± 0.02 |

The F1 score comparison is also displayed in **Figure 3a.** The MF detection scores of OMG-Net are significantly higher ($p < 0.001$) in all three types of human tumours within the testing set of MIDOG++. **Figure 3b** shows the benefit of combining multi-centre data, as the increased number of MFs for training correlated with an increase in the holdout F1-score.

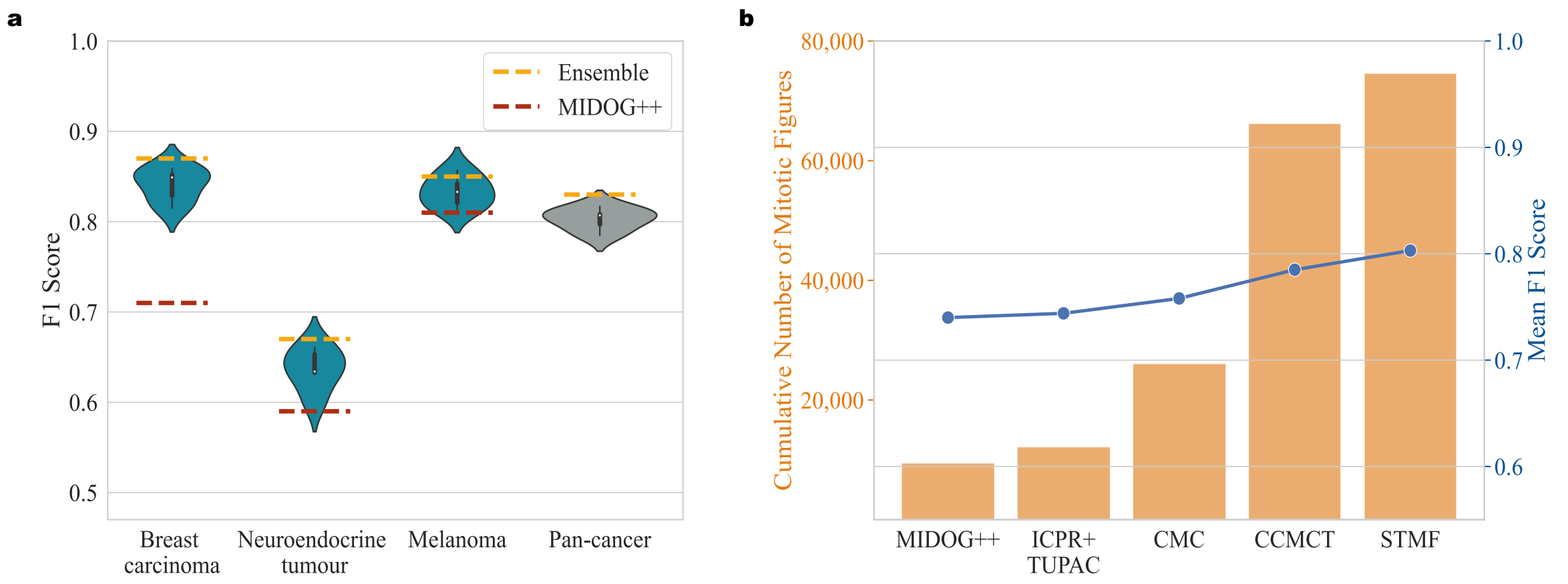

**Figure 3: Detection performance. a** The testing F1 scores of the proposed framework, where the yellow dashed lines mark the ensemble F1 scores and the red dashed lines mark the mean F1 scores reported by MIDOG++. **b** The changes in the average F1 score as more mitotic figures (MFs) are included in training.

### *Reviewing the SAM masks enhanced the detection performance*

As shown in **Figure 4a,** the appearance of MFs is highly diverse, exemplified by atypical MFs. The segmented mask may not fully cover the MFs or may contain background noise. To refine the training process, we reviewed the masks in the human subset of MIDOG++ (4435 in breast carcinoma, 2075 in melanoma and 2400 in neuroendocrine tumour), and adjusted the SAM prompt when required. The impact of this manual curation was assessed by comparing the F1 scores of the models only with RGB images (RGB Classifier), the score of the model with zero-shot SAM mask input (RGB-M0 Classifier) as well as the score of the model with reviewed and refined masks (RGB-M1 Classifier), with results shown in **Figure 4b.**

Compared to the model without masks (RGB), the RGB-M0 model yielded higher F1 scores for detecting MFs from breast carcinoma ($p = 0.011$) and melanoma ($p = 0.001$) but not for neuroendocrine tumours. Upon further analysis, we noted that the fraction of masks requiring a second adjustment was higher in neuroendocrine tumours (16%), compared to breast carcinoma (8%) and melanoma (5%). As predicted, the RGB-M1 Classifier showed the best performance and significantly outperformed the RGB Classifier for breast carcinoma ($p < 0.001$), melanoma ($p < 0.001$) and neuroendocrine tumours ($p = 0.021$). We conclude that the low-quality masks, which may include surrounding backgrounds or exclude part of the nuclei (**Figure 4b**), can impact the performance of the RGB-M0. Further examples of failed prompts are shown in **Supplementary Figure 1**.

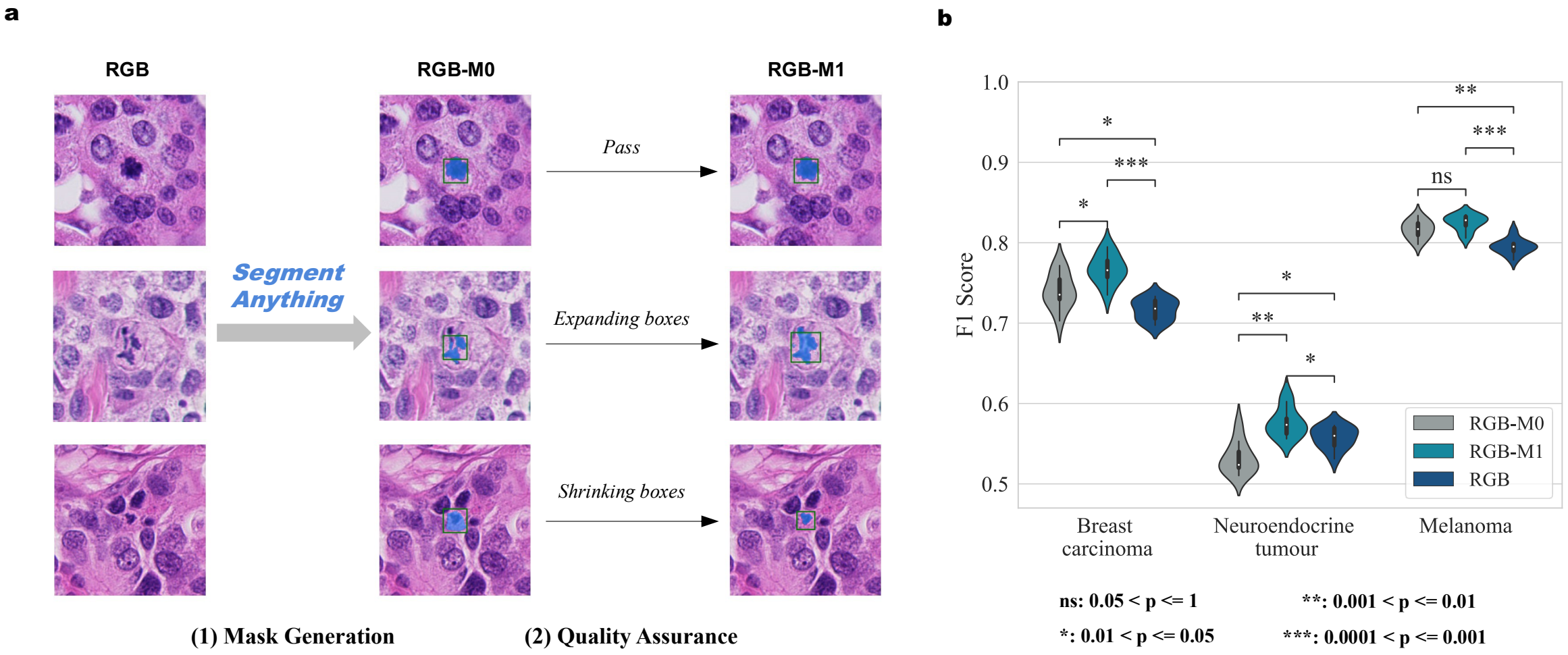

**Figure 4: The contribution of Segment Anything (SAM) masks. a** Illustration of the quality assurance process for MIDOG++ human subset. **b** F1 scores of the classifier using only RGB images (RGB Classifier), the classifier using additional SAM masks (RGB-M0 Classifier), and the model using reviewed SAM masks (RGB-M1 Classifier).

### *Canine mitotic figures help to train the detection of human mitotic figures*

The merged and uniform dataset contains a significant proportion of canine MFs with examples from both human and canine WSI displayed in **Figure 5a**. The inclusion of the canine data significantly improved the detection of MFs in breast carcinoma ($p = 0.007$) and neuroendocrine tumours ($p = 0.015$) and the F1 score in melanoma was also marginally increased ($p = 0.080$) **(Figure 5b)**.

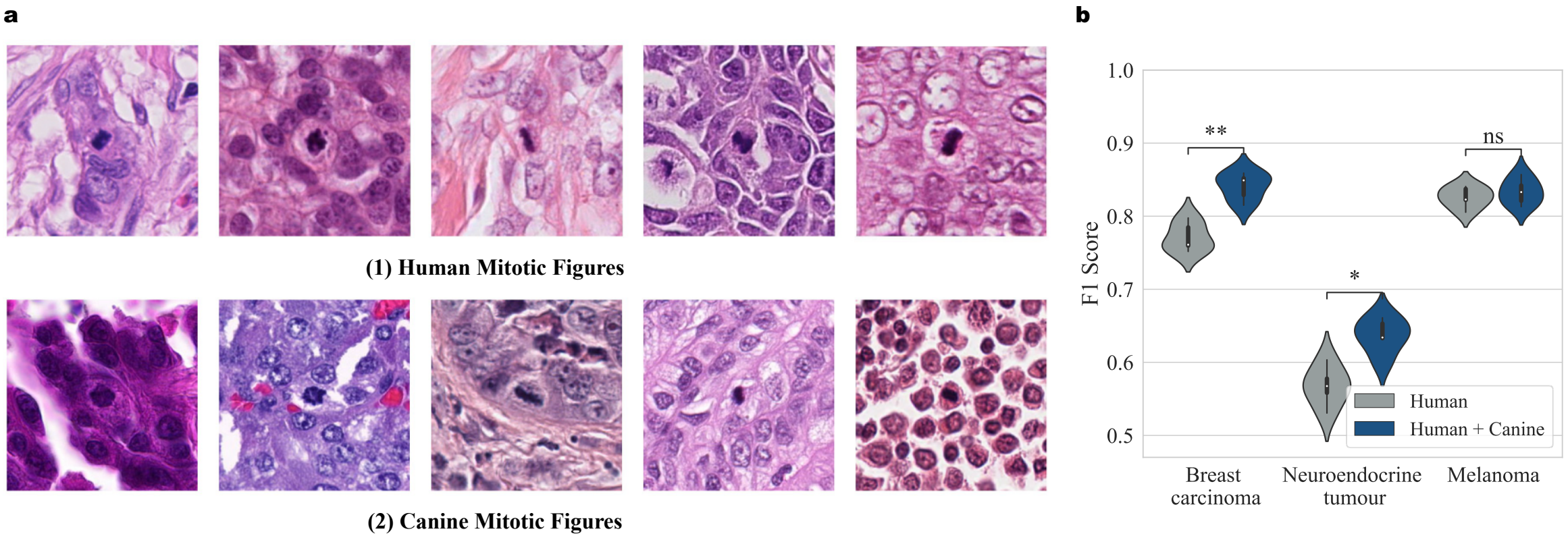

**Figure 5: Including the canine mitotic figures (MFs) for training improves the detection. a** Example of MFs in human and canine haematoxylin and eosin (H&E)-stained sections. **b** The F1 scores of the models trained with only human data and with both human and canine data.

### *Including mitotic-like figures and non-mitotic objects is key to improving model precision*

Besides MFs, MLFs were also labelled in the original dataset. MLFs represent morphological structures that resemble MFs including pyknotic nuclei, apoptotic bodies, and neutrophil polymorph amongst others, often misclassified as MFs. An example is displayed in **Figure 6a**. Apart from MLFs and tumour cells, the SAM-curated dataset contains other cells, including immune cells, red blood cells and any objects at the cell level such as artefacts, segmented by the SAM during data curation to present the classifier with a heterogeneous set of data. These were used in the training step to augment the original dataset and provide the model with a diverse representation of segmented objects. **Figure 6b** shows that the model including non-MF objects (SAM-AUG) has significantly higher precision for all three types of tumours (p = 0.008) compared to the model trained only with MFs and MLFs (Original). As expected, the recall remains unchanged, and the overall F1 scores are improved (p = 0.007).

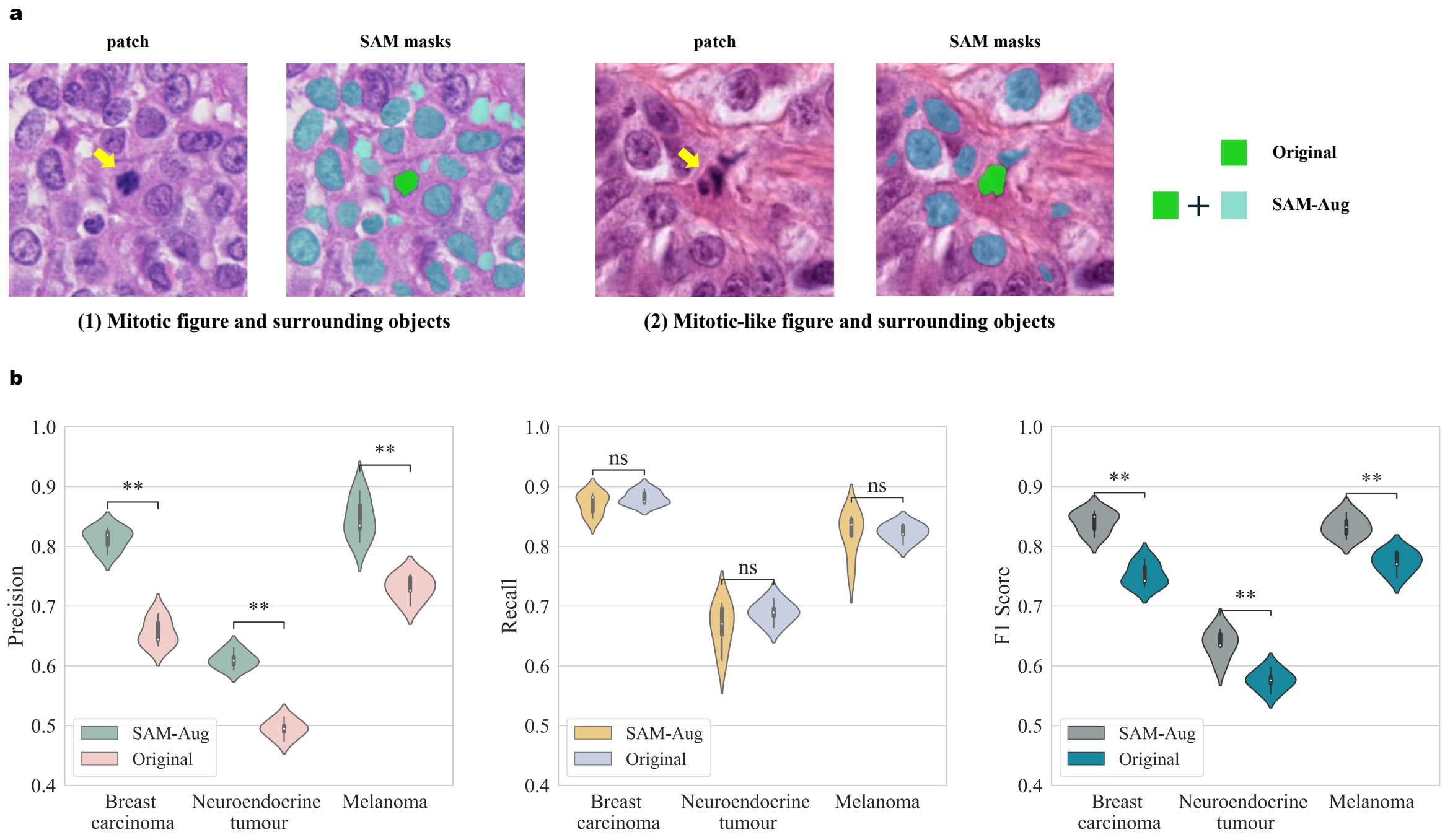

**Figure 6: Including mitotic-like figures (MLFs) and non-mitotic objects for training improves the detection. a** Example of patches containing a mitotic figure (MF) (left) and a MLF (right). The MFs and MLFs are masked in green (Original data). The surrounding cellular components segmented by Segment Anything (SAM) are marked in light blue and are added to the MFs and MLFs (SAM-Aug data). **b** The precision, recall and F1 scores of the model trained with the Original data and the model trained with SAM-Aug data.

## Discussion

### *Nuclei contours represent a key feature for improving mitotic-figure detection*

In this study, OMG-Net showed significantly improved MF detection performance in all three types of tumours compared with the testing set of the current state-of-the-art MIDOG++. This improvement was achieved both by using a larger, multi-source, SAM-enhanced dataset of MFs (**Figure 1**) and by integrating within the network the mask of the MFs' nuclei (**Figure 4b)**.

The accuracy of our models varied considerably across different tumour types, with neuroendocrine tumours exhibiting significantly lower performance, which was consistent with the results of the MIDOG++ algorithm. In parallel, we observed a higher proportion of low-quality masks in neuroendocrine tumours (16%) compared to breast carcinoma (8%) and melanoma (5%), suggesting that the quality of the training data may have contributed to the disparities in model performance across these cancer types. Even then, manual curation of the masks helped improve significantly the model detection performance.

To decide which foundational model to select as a nuclei detector, we evaluated published fine-tuned variants of SAM against the overall mask quality for cells in histology images. Specifically, we tested MedSAM (Ma, et al., 2024), which was fine-tuned on multiple medical image modalities, and CellSAM (Israel, et al., 2023), which was fine-tuned on microscopy images. By inspecting the number of cells detected and the layout of the masks produced by both algorithms, we concluded that the quality of masks was degraded in both models and included a higher proportion of omitted cells (**Supplementary Figure 2**). We attribute the reduced performance of the fine-tuned SAM models to different architectures used compared to the original SAM model. OMG-Net uses the highest capacity SAM variant (ViT-H), whereas MedSAM and CellSAM are fine-tuned on the lighter ViT-B, which may lead to reduced performance due to limited model capacity. Based on this analysis, we elected to keep the original SAM as the cell detector in our study. Future work will include refining the SAM object-proposal method for H&E-stained specific cell types

***Decoupling segmentation and classification helps improve the detection performance***

Object detection models such as Faster R-CNN (Ren, et al., 2016), RetinaNet (Lin, et al., 2018) and YOLO (Redmon, et al., 2016) have been widely used for MF detection (Mahmood, et al., 2020) (Bertram, et al., 2021). These models integrate in a single model an object proposal network with a primary classifier. However, these models suffer from the imbalanced loss problem, as the cell segmentation and classification loss have inherently unequal magnitudes. The gradient updates that occur during backpropagation can be dominated by the loss function with the larger norm (Chen, et al., 2018), leading to suboptimal training and convergence issues. This becomes even more prominent when dealing with small datasets or complex objects, as the imbalance in the loss functions' impact can significantly hinder the model's ability to learn effectively from the limited available data (Argyriou, et al., 2006). The use of integrated object detection models in histopathological studies has been shown to generate false positive results due to the complex and variable nature of cell morphology.

More recently, it has been demonstrated (Çayır, et al., 2022; Sohail, et al., 2021) that integrating a secondary classifier, trained on MFs and other objects such as MLFs, to review and reject false positive cases improves a framework's precision. This approach limits the imbalanced loss problem, as the segmentation loss is excluded in training the additional classifiers. However, these methods add unnecessary complexity to the network since two classifiers must be trained.

To mitigate the imbalanced loss problem, we elected to separate entirely the object detection and classification steps. This offers an innovative approach that differs from those previously published. Instead of training an object detection model for generating objects that are highly likely to be MFs, all the objects at the cell scale are segmented by SAM from the ROIs and classified, improving the sensitivity of our model. Other objects, including immune cells, cells not in mitosis, and artefacts generated during the data preparation stage, can also be used to train the classifier, improving its capability to reject false positives.

***Large-scale MF datasets provide a resource for the development of pan-cancer models***

Large-scale datasets are crucial for developing AI models capable of detecting MFs effectively in a variety of cancer types and overcoming the challenges posed by the heterogeneity of staining and scanning protocols. Here, we propose a workflow for creating a reliable MF dataset:

1) H&E destaining and employing immunohistochemistry for enhanced detection: efficient generation of a large-scale image dataset with accurate labels by detecting a substantial number of MFs on WSIs.
2) Continuous Data Curation: improve data quality by employing Segment Anything (SAM) to delineate precisely mitotic figure (MF) nuclei, followed by meticulous manual refinement of the generated contours.
3) Active learning: iteratively train and refine the model using a pathologist-in-the-loop approach, enabling efficient review of detected mitotic figures (MFs), and incorporating Mitotic-Like Figures (MLFs) and non-mitotic objects into the database for enhanced model performance

These steps are required as it is not feasible for pathologists to annotate MFs in the numbers and the precision required by AI models, thereby affecting the diversity and size of the dataset and, consequently, the detection accuracy of the trained model. Nevertheless, each of these steps encounters limitations.

Performing immunohistochemistry following the destaining procedure of H&E-stained sections allows for the rapid and largely specific detection of MFs (specificity >99%) (Kim, et al., 2017). Still, it is not a perfect process as cells in the G2 phase of the cell cycle can exhibit weak immunoreactivity (Tacha, 2015) as well as being prone to false-negative immunoreactivity due to the age of the slide and fixation method (Hendzel, et al., 1997). This restaining procedure also does not detect MLFs, which is crucial to enhance the model specificity.

Active learning can help identify MLFs but a consensus view of MF/MLF cannot always be reached by pathologists. This study highlighted the acknowledged problem of interobserver variation of MF by pathologists (Veta, et al., 2016; Robbins, et al., 1995) which is compounded when interpreting MFs on digitised slides as it is not possible to adjust the focus plane on cells of interest. During our revision process, a notable proportion (13.8%) of AI-detected cells were categorised as "equivocal" (**Supplementary Figure 3**). A secondary review of these images performed by at least two experienced pathologists resolved some of these images but differences in opinion remained in 9.5% of AI-detected MFs.

Finally, despite the limitations discussed above, the integration of immunohistochemistry for MF detection following the destaining of H&E sections, data curation, active learning, and consensus-based review by experienced pathologists enabled us to mitigate the challenges in creating a large-scale database and developing an improved, pan-cancer MF detection model.

## Conclusion

We have established a large-scale MF dataset by integrating five open-source datasets acquired from multiple centres including an in-house dataset of STT. Using the curated dataset, we employed a novel two-step framework, OMG-Net, where SAM served as the object detector followed by an adapted ResNet18 as the MF classifier. This approach improved the accuracy of MF detection from various human tumours including breast carcinoma, neuroendocrine tumours and melanoma compared to existing state-of-the-art models. Future steps include a head-to-head prospective assessment of this model with pathologists' scores for MFs before introduction into safe clinical practice.

## Data availability

All MF images provided by the open-source datasets and their SAM-dilated contours are available without restriction via Zenodo in accordance with the UKRI Common principles on research data. The MF images in STMF are available upon reasonable request.

## Code availability

All code to reproduce the results, when coupled with the dataset available on Zenodo, is provided free of use at https://github.com/cacof1/DigitalPathologyAI.

## Author contributions: CRediT

**Zhuoyan Shen**: Conceptualization, Formal analysis, Investigation, Methodology, Validation, Visualization, Writing – original draft, Writing – review and editing. **Mikael Simard**: Methodology, Writing – review and editing. **Douglas Brand**: Investigation, Writing – review and editing. **Vanghelita Andrei**: Data curation, Investigation. **Ali Al-Khader**: Data curation, Investigation. **Fatine Oumlil**: Data curation, Investigation. **Katherine Trevers**: Project administration. **Thomas Butters**: Data curation, Investigation. **Simon Haefliger**: Data curation, Investigation. **Eleanna Kara**: Data curation, Investigation. **Fernanda Amary**: Resources, Data curation, Investigation. **Roberto Tirabosco**: Resources, Data curation, Investigation. **Paul Cool**: Resources, Investigation, Writing – review and editing. **Gary Royle**: Supervision, Investigation, Writing – review and editing. **Maria A. Hawkins**: Supervision, Writing – review and editing. **Adrienne M. Flanagan**: Conceptualization, Funding acquisition, Supervision, Resources, Data curation, Investigation, Writing – review and editing. **Charles-Antoine Collins Fekete**: Conceptualization, Funding acquisition, Supervision, Resources, Data curation, Investigation, Methodology, Writing – review and editing.

## Acknowledgements

This project is supported by the UKRI Future Leaders Fellowship (MR/T040785/1), EPSRC Research Grant NR1 (EP/Y020030/1), the Radiation Research Unit at the Cancer Research UK City of London Centre Award (C7893/A28990), as well Sarcoma UK (Award SUK18.2021). AMF and KT are supported by the National Institute for Health Research, UCLH Biomedical Research Centre, and the CRUK Experimental Cancer Centre as well as the Royal National Orthopaedic Hospital R& D Department. The Royal National Orthopaedic Hospital, Stanmore, Middlesex HA7 4LP. SH was funded by the Children's Cancer Foundation Basel (grant: C23-2021-21). TB is a PhD Clinical Fellow, funded by the Jean Shanks Foundation and the Pathological Society of Great Britain and Ireland. EK was funded by the Royal National Orthopaedic Hospital.

## Declaration of competing interest

The authors declare no competing interests.

# Supplementary Materials

**Supplementary Figure 1: Low-quality masks.** Examples of mitotic figure (MF) masks (the blue masks) that are badly delineated by Segment Anything (SAM). The MFs are indicated by yellow arrows, the green boxes are the prompts used when deploying the SAM.

**Supplementary Figure 2: Cell detection performance of the original and fine-tuned Segment Anything (SAM).** Objects segmented by the original SAM, MedSAM and CellSAM on three example patches. N is the number of objects detected in each patch.

**Supplementary Figure 3: Figures with disagreement between pathologists.** The figures indicated by the yellow arrows were the examples labelled as "equivocal" by the junior pathologists and were sent to senior pathologists for secondary annotating.

**Supplementary Table 1.** Summary of datasets and scanners.

**Supplementary Table 2.** Number of whole slide images (WSIs) and mitotic figures (MFs) for each diagnosis in STMF.

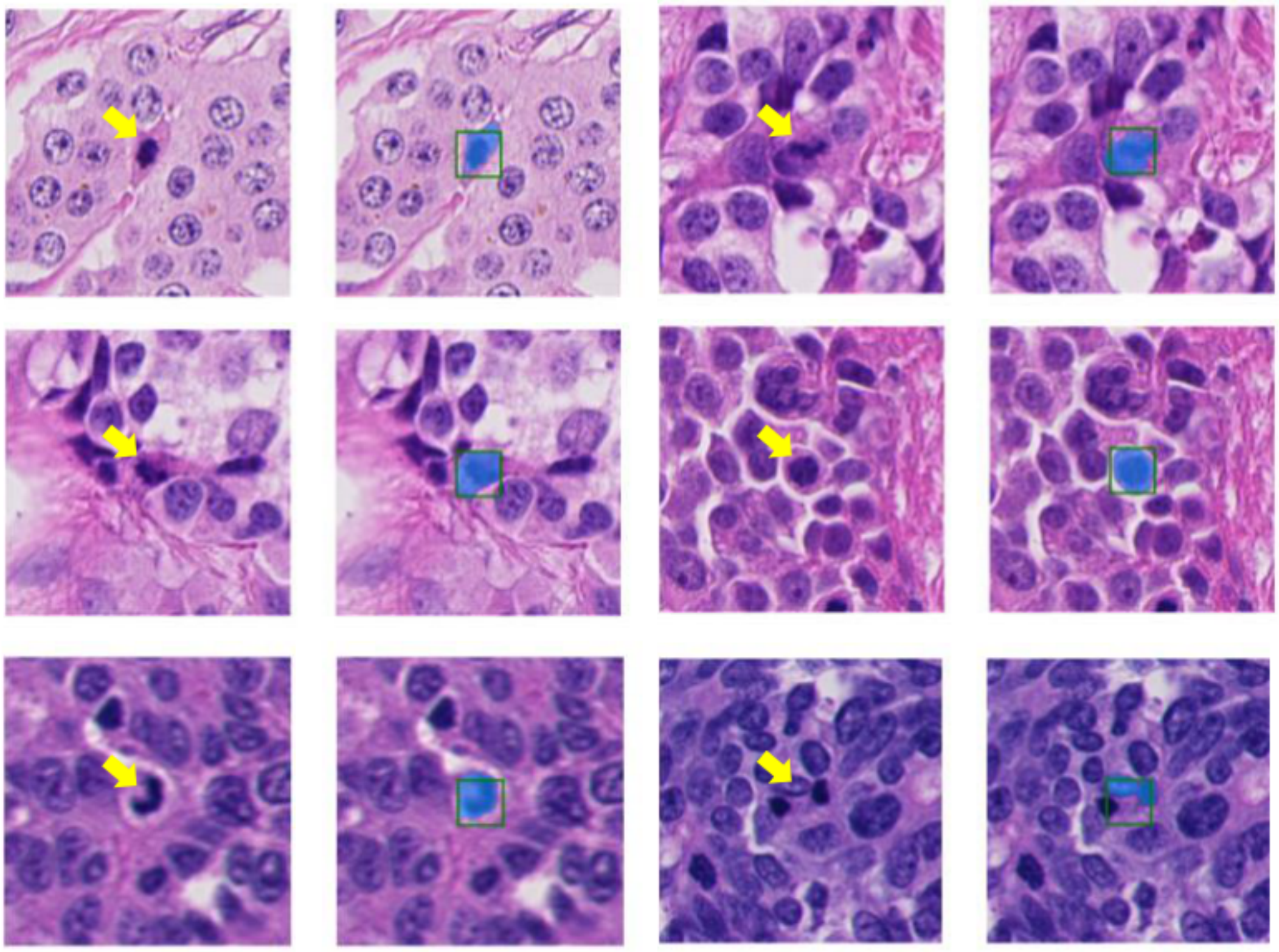

**Supplementary Figure 1: Low-quality masks.** Examples of mitotic figure (MF) masks (the blue masks) that are badly delineated by Segment Anything (SAM). The MFs are indicated by yellow arrows, the green boxes are the prompts used when deploying the SAM.

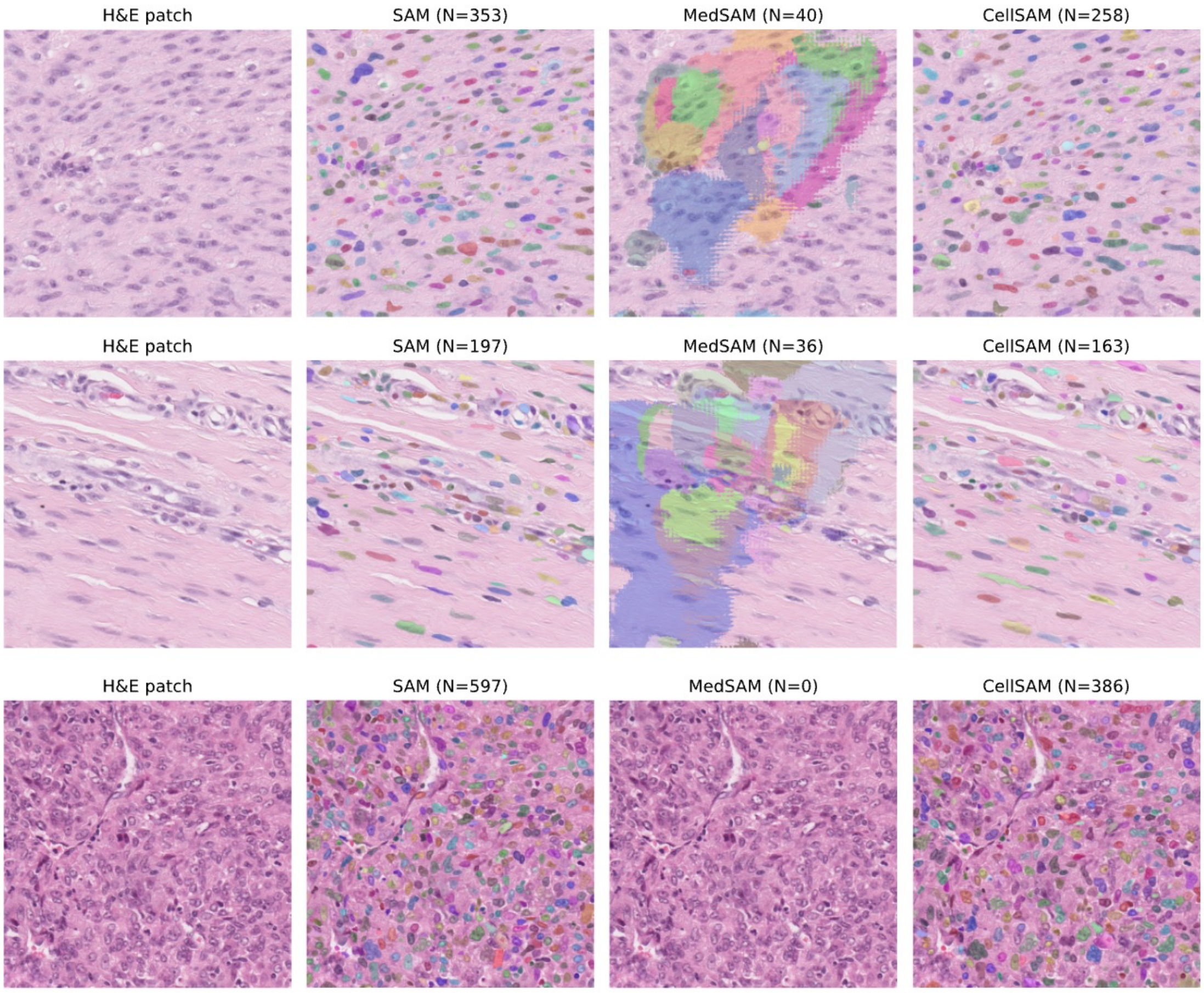

**Supplementary Figure 2: Cell detection performance of the original and fine-tuned Segment Anything (SAM).** Objects segmented by the original SAM, MedSAM and CellSAM on three example patches. N is the number of objects detected in each patch.

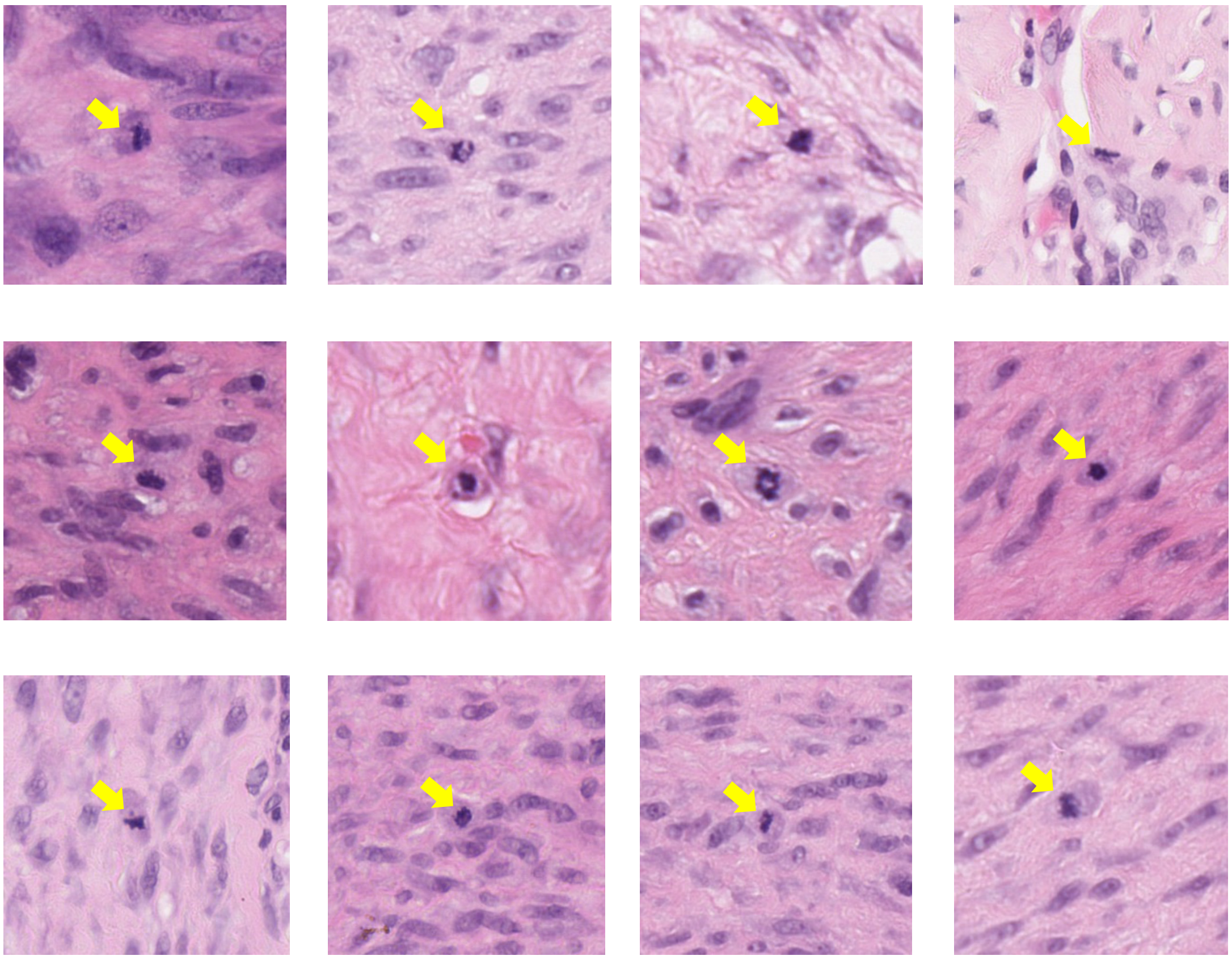

**Supplementary Figure 3: *F*igures with disagreement between pathologists.** The figures indicated by the yellow arrows were the examples labelled as “equivocal” by the junior pathologists and were sent to senior pathologists for secondary annotating.

**Supplementary Table 1. Summary of datasets and scanners**

| Dataset | Scanners |
|---|---|
| ICPR | Aperio Scanscope XT<br>Hamamatsu Nanozoomer 2.0-HT |
| TUPAC | Leica SCN400 |
| MIDOG++ | Hamamatsu XR |
| | Hamamatsu S360<br>3DHistech Pannoramic Scan II<br>Aperio ScanScope CS2 |
| CMC | Aperio ScanScope CS2 |
| CCMCT | Aperio ScanScope CS2 |
| STMF | Aperio ScanScope CS2 |

**Supplementary Table 2. Number of whole slide images (WSIs) and mitotic figures (MFs) for each diagnosis in STMF.**

| **Diagnosis** | **Number of WSIs** | **Number of MFs** |
|---|---|---|
| Angiosarcoma | 4 | 672 |
| Chondrosarcoma | 3 | 105 |
| Ewing Sarcoma | 3 | 54 |
| Giant Cell Tumour of bone | 3 | 195 |
| Leiomyosarcoma | 12 | 381 |
| Solitary Fibrous Tumour | 99* | 652 |
| Dedifferentiated liposarcoma | 1 | 54 |
| Melanoma | 3 | 411 |
| Myeloma | 1 | 44 |
| Myxofibrosarcoma | 10 | 826 |
| Nerve Sheath Tumour | 1 | 49 |
| MPNST | 1 | 88 |
| Osteosarcoma | 9 | 130 |
| Pleomorphic Sarcoma | 30 | 2228 |
| Rhabdomyosarcoma | 3 | 66 |
| NFATC2-Sarcoma | 1 | 43 |
| Spindle Cell Sarcoma | 10 | 1326 |
| Synovial Sarcoma | 7 | 502 |
| Desmoid fibromatosis | 74* | 34 |
| Superficial fibromatosis | 53* | 37 |

*Annotated by active learning. The other data is produced by pHH3 staining.